\documentclass{article}

\usepackage{tabularx}
\usepackage{PRIMEarxiv}
\usepackage{amsmath}
\usepackage{algorithm}
\usepackage{algorithmic}
\usepackage[utf8]{inputenc} 
\usepackage[T1]{fontenc}    
\usepackage{hyperref}       
\usepackage{url}            
\usepackage{booktabs}       
\usepackage{amsfonts}       
\usepackage{nicefrac}       
\usepackage{microtype}      
\usepackage{lipsum}
\usepackage{fancyhdr}       
\usepackage{graphicx}       
\graphicspath{{media/}}     

\title{CATFormer: When Continual Learning Meets Spiking Transformers With Dynamic Thresholds
\thanks{This version has been accepted for publication in the proceedings of the Neuro for AI \& AI for Neuro Workshop at AAAI 2026 (PMLR).} 
}

\author{
  Vaishnavi Nagabhushana\\
  SustainAI Lab, MFSDS\&AI\\
  IIT Guwahati\\
  \texttt{n.vaishnavi@iitg.ac.in} \\
    \And
  Kartikay Agrawal\\
  SustainAI Lab, MFSDS\&AI\\
  IIT Guwahati\\
  \texttt{a.kartikay@iitg.ac.in} \\
     \And
  Ayon Borthakur\\
  SustainAI Lab, MFSDS\&AI\\
  IIT Guwahati\\
  \texttt{ayon.borthakur@iitg.ac.in}
}

\begin{document}
\maketitle
\begin{abstract}
Although deep neural networks perform extremely well in controlled environments, they fail in real-world scenarios where data isn't available all at once, and the model must adapt to a new data distribution that may or may not follow the initial distribution. Previously acquired knowledge is lost during subsequent updates based on new data. a phenomenon commonly known as catastrophic forgetting. In contrast, the brain can learn without such catastrophic forgetting, irrespective of the number of tasks it encounters. Existing spiking neural networks (SNNs) for class-incremental learning (CIL) suffer a sharp performance drop as tasks accumulate. We here introduce CATFormer (Context Adaptive Threshold Transformer), a scalable framework that overcomes this limitation. We observe that the key to preventing forgetting in SNNs lies not only in synaptic plasticity but also in modulating neuronal excitability. At the core of CATFormer is the Dynamic Threshold Leaky Integrate-and-Fire (DTLIF) neuron model, which leverages context-adaptive thresholds as the primary mechanism for knowledge retention. This is paired with a Gated Dynamic Head Selection (G-DHS) mechanism for task-agnostic inference. Extensive evaluation on both static (CIFAR-10/100/Tiny-ImageNet) and neuromorphic (CIFAR10-DVS/SHD) datasets reveals that CATFormer outperforms existing rehearsal-free CIL algorithms across various task splits, establishing it as an ideal architecture for energy-efficient, true-class incremental learning. 
\end{abstract}
\section{Introduction}

Progress in physical AI holds immense promise for enhancing real-world capabilities across robotics, near-sensor edge devices, and autonomous systems. A critical challenge on these platforms is learning and predicting cyclically across extended deployments with minimal resource utilisation.  Model updates are often essential due to sequentially arriving data and distributional shifts \cite{chaudhry2018efficient, wang2024comprehensive}. But naively training standard deep neural networks (MLPs, CNNs, or even modern transformers) from scratch repeatedly typically results in \emph{catastrophic forgetting} of previously acquired knowledge. In battery-operated, memory or bandwidth-constrained physical agents, data rehearsal (i.e., storage and replay of past data during training on new data) is often infeasible due to energy, onboard memory, privacy, or regulatory constraints \cite{lesort2020continual}.

Energy-efficient learning architectures are essential in these applications, and \textbf{Spiking neural networks (SNNs)} have become a well-established solution, offering event-driven sparse computations. Recent advances have brought class-incremental learning (CIL) to SNNs with early efforts mainly on small-scale tasks (e.g., CIFAR-10) and, more recently, on larger, more realistic benchmarks like CIFAR-100 using various incremental task regimes \cite{ni2025alade, dsdsnn}. However, prior efforts in SNN-based CL have predominantly relied on convolutional (CNN) architectures. Recently, transformers and their variants have dominated performance in numerous AI applications. Vision transformers \cite{dosovitskiy2020image} can also capture global dependencies and contextual understanding. And, SpikFormer \cite{zhou2023spikformer} extends the standard vision transformer paradigm to SNNs by combining its strengths with energy efficiency. Yet, its potential for continual learning remains largely untapped. Hence, we here design CATFormer, a SNN-based transformer for long class incremental learning. 
\begin{figure*}[!htbp]
    \centering
    \includegraphics[width= 0.89\textwidth]{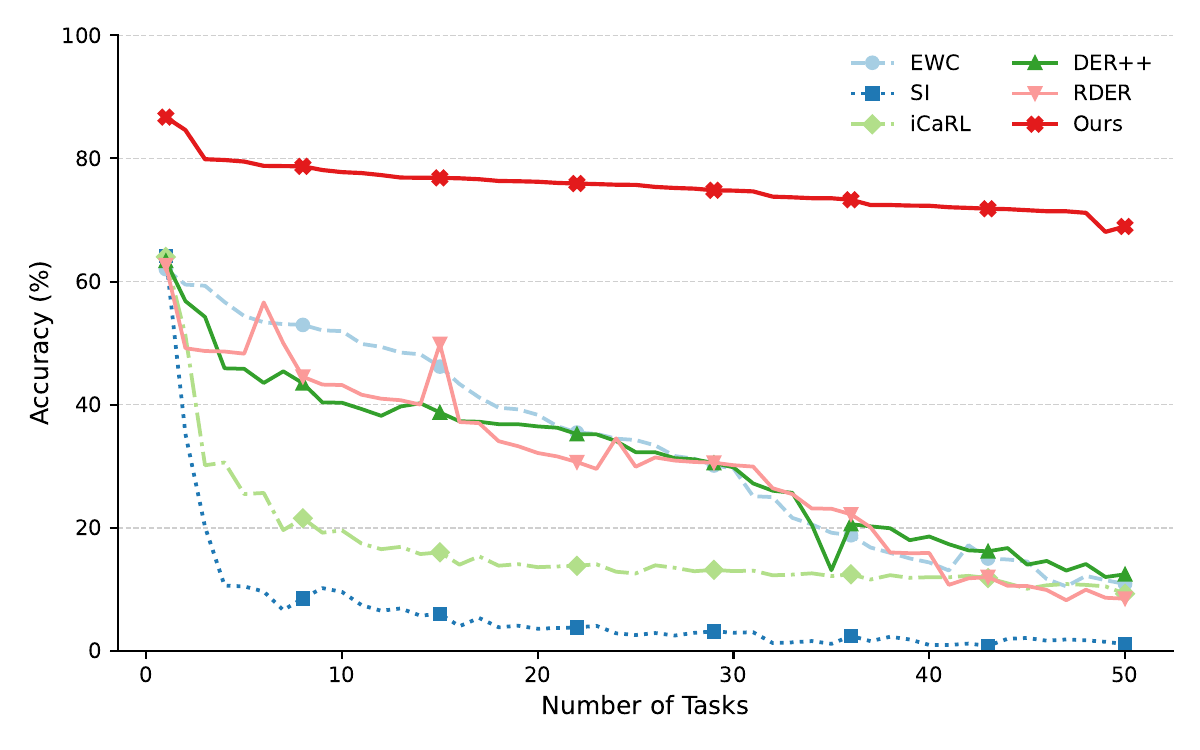}
    \caption{Test performance variation with respect to the progress in the number of trained tasks (for a maximum of 50 tasks). CATFormer (ours) maintains consistent performance with other existing CIL methods when implemented on a Spiking Transformer. All methods are evaluated on CIFAR 100.}
    \label{fig: cifar100 on 50 tasks}
\end{figure*}
CATFormer is inspired by the brain, where resistance to forgetting has been proposed to be closely linked to neuromodulation \cite{masse2018alleviatin, beaulieu2020learning}. Neuromodulators such as acetylcholine, dopamine, serotonin, and norepinephrine mediate changes in neural circuit behaviour by altering plasticity or excitability. For instance, acetylcholine plays a central role by modulating membrane excitability and synaptic plasticity in hippocampal and cortical networks, thereby enabling the rapid encoding of new memories while transiently lowering neuronal firing thresholds to suppress interference from prior information \cite{hasselmo1995cholinergic, grossberg2017acetylcholine}. Similar cholinergic modulation of plasticity is also observed in rodent piriform cortex \cite{hasselmo2006role}. These processes in the brain dynamically regulate neuronal excitability and firing thresholds \cite{liu2022biologically,oh2015increased}, enabling selective pathway activation and memory consolidation while preventing interference \cite{xu2005activity,farmer2012learning}. Computational models and experimental studies further demonstrate that neuromodulators broadly demonstrate task and context-specific routing of information by reshaping network dynamics \cite{tsuda2026neuromodulators, masse2018alleviatin,hammouamri2022mitigating}. Although we don't claim that CATFormer is completely biologically plausible, these multi-scale neuromodulatory effects inspire our approach to implement adaptive thresholds within CATFormer, serving as an analogy to support plasticity in exemplar-free continual learning.

In this work, we systematically study class incremental learning in a spiking vision transformer \cite{zhou2023spikformer}. We analyse how biologically inspired, dynamic spiking thresholds influence continual learning and propose mechanisms that foster robustness across encountered tasks. This is crucial for \emph{physical AI and robotics}: real-world robots and edge agents must adapt over months or years, often encountering dozens or even hundreds of different skills or operating regimes \cite{lesort2020continual}. Thus, we rigorously evaluate our approach at unprecedented scales, including challenging 50 and 100-task sequences, providing a rigorous testbed for long-term continual learning relevant to autonomous robotics and physical edge applications. Moreover, our data rehearsal-free protocol ensures that results directly reflect core algorithmic advances, rather than storage-based workarounds. This can be observed in Figure \ref{fig: cifar100 on 50 tasks}, where the model maintains its accuracy across longer task sequences. We observe a phenomenon we call \textit{reverse forgetting},  where the model actually learns more effectively when exposed to fewer classes per task. This scenario is more reflective of real-world settings—for example, in robotics or lifelong learning, the model is unlikely to encounter 20 or 50 new classes all at once \cite{lesort2020continual, masse2018alleviatin}.

\textbf{We make the following advances in this work:}

\begin{itemize}
    \item To the best of our knowledge, we present the first CIL framework designed for spiking vision transformers, closing a major gap in the field.
    \item We propose a novel, biologically-inspired adaptation mechanism where a frozen backbone learns new tasks primarily through task-specific dynamic thresholds. This approach serves as the core mechanism for preventing catastrophic forgetting without requiring network growth or storing raw data exemplars.
    \item We demonstrate state-of-the-art performance among exemplar-free SNNs and, critically, show that CATFormer is uniquely scalable to long task sequences. Unlike prior methods that degrade, our model maintains, or even improves, its accuracy on challenging benchmarks comprising up to 100 incremental tasks, i.e., exhibiting a \textbf{reverse forgetting} trend.
\end{itemize}

\section{Related Work}

\subsection{Continual Learning Paradigms}
Continual learning methods often fall into three main approaches \cite{threeparadiagms}. \textbf{Regularisation-based} methods mitigate forgetting by constraining updates to parameters deemed important for previous tasks, thus preventing overwriting of past knowledge \cite{ewc,zenke2017continual}. \textbf{Rehearsal-based} methods maintain a buffer of stored previous data samples either as original images or as representations for replay during training, improving memory retention but incurring increased storage and privacy concerns \cite{rebuffi2017icarl,der++,representationexpansion,cvpr25mem_eff_reh}. \textbf{Architecture-based} methods adapt the network structure dynamically\cite{dsdsnn,selfexpansioncvpr}, such as by adding modules or selecting task-specific sub-networks, balancing plasticity and stability at the cost of computational overhead \cite{rusu2016progressive,fernando2017pathnet}. While these architecture-based continual learning methods are effective, they have scalability limitations and memory overhead as the task count increases. 

In the brain, context-dependent signals from regions like the prefrontal cortex project across cortical areas, allowing neural circuits to adaptively process information based on the task at hand
\cite{engel2001dynamic_pnas,miller2001integrative_pnas}. Previous techniques leveraged EWC \cite{ewc} with a gating mechanism to stabilise training for feedforward and recurrent architectures \cite{masse2018alleviatin}.

\subsection{Transformers in Continual Learning}
Transformers, with their self-attention mechanisms, have recently emerged as a powerful alternative to convolutional neural networks (CNNs) for continual learning due to their ability to model long-range dependencies and the ease with which pretrained architectures can be extended \cite{selfexpansioncvpr,liang2024inflora}. Recent work has focused on parameter-efficient fine-tuning techniques like Low-Rank Adaptation (LoRA) \cite{he2025cllora,liang2024inflora} to update large pre-trained transformers continually with limited overhead \cite{selfexpansioncvpr,cvpr25taskspecificadapters}. However, these standard vision transformers require attention computations, which in turn lead to energy inefficiency due to their heavy matrix multiplications. Hence, we move towards spiking vision transformers (3.31 times more energy efficient) \cite{zhou2023spikformer}, which can eliminate these heavy computations. Continual learning research on spiking vision transformers remains unexplored. Our work presents a data rehearsal-free continual learning on spiking vision transformers trained from scratch, addressing these challenges and expanding the landscape of SNN continual learning.

\subsection{Spiking Neural Networks for Continual Learning and Neuromodulation Inspiration}
Early SNN continual learning methods adapted classical regularisation and rehearsal methods that are limited to CNNs \cite{dsdsnn,lin2025onlinecontinuallearningspiking}. Recent work \cite{ni2025alade} shows significant improvement by incorporating rehearsal buffers into a method inspired by DER++ \cite{der++} to enhance performance, but it violates data privacy and is memory inefficient. Preliminary work on converting the idea of neuromodulation to the circuit level was demonstrated by \cite{hammouamri2022mitigating}, where the thresholds of the current layer were modulated based on the previous layer's excitation. Our model introduces adaptive, task-specific dynamic thresholds. This mimics neuromodulatory circuits that release context signals, modulating downstream neurons' responses, corresponding to our task-ID routing mechanism that dynamically tunes neuron thresholds per task to enable data-replay-free continual adaptation.

\subsection{Dynamic Thresholds in Spiking Neural Networks}

Adaptive firing thresholds have been shown to improve temporal precision and robustness in SNNs \cite{ding2022biologically,huang2016adaptive,wei2023temporal}. Although many mechanisms have been extensively studied in neuroscience, only a handful of works have investigated bioinspired dynamic threshold rules to improve SNN generalisation. Existing approaches include a dynamic threshold update rule that adaptively scales firing thresholds to prevent excessive activity \ cite {hao2020biologically}, double exponential functions for threshold decay \cite{shaban2021adaptive}, and predefined target firing counts \cite{kim2021spiking}. BDETT computes thresholds via average membrane potentials for neuronal homeostasis \cite{ding2022biologically}, but excessive spiking from highly active neurons reduces sparsity.
However, prior work has predominantly focused on single-task adaptation rather than task-specific thresholding for continual learning. By freezing synaptic weights after the initial task and relying solely on learnable, per-task thresholds for plasticity, our approach introduces a novel mechanism to prevent catastrophic forgetting in a rehearsal-free framework. Together with lightweight task gating, this enables stable continual learning across up to 100 tasks on static and neuromorphic datasets, setting a new standard in memory and energy-efficient SNN continual learning.

\section{Methodology}
We introduce a data rehearsal-free framework for class-incremental learning (CIL) in Spiking Neural Networks (SNNs) that robustly accommodates new classes over time without catastrophic forgetting. Unlike existing systems, our entire system is trained without a separate exemplar representation buffer from previous learning across tasks \cite{ni2025alade,der++}. The proposed architecture leverages two key innovations: \textit{task-specific (Context Adaptive) dynamic neuronal thresholds} and a \textit{gated inference mechanism}, combined through a two-stage training protocol.

\begin{figure*}[!htbp]
\centering
\includegraphics[width=0.85\textwidth]{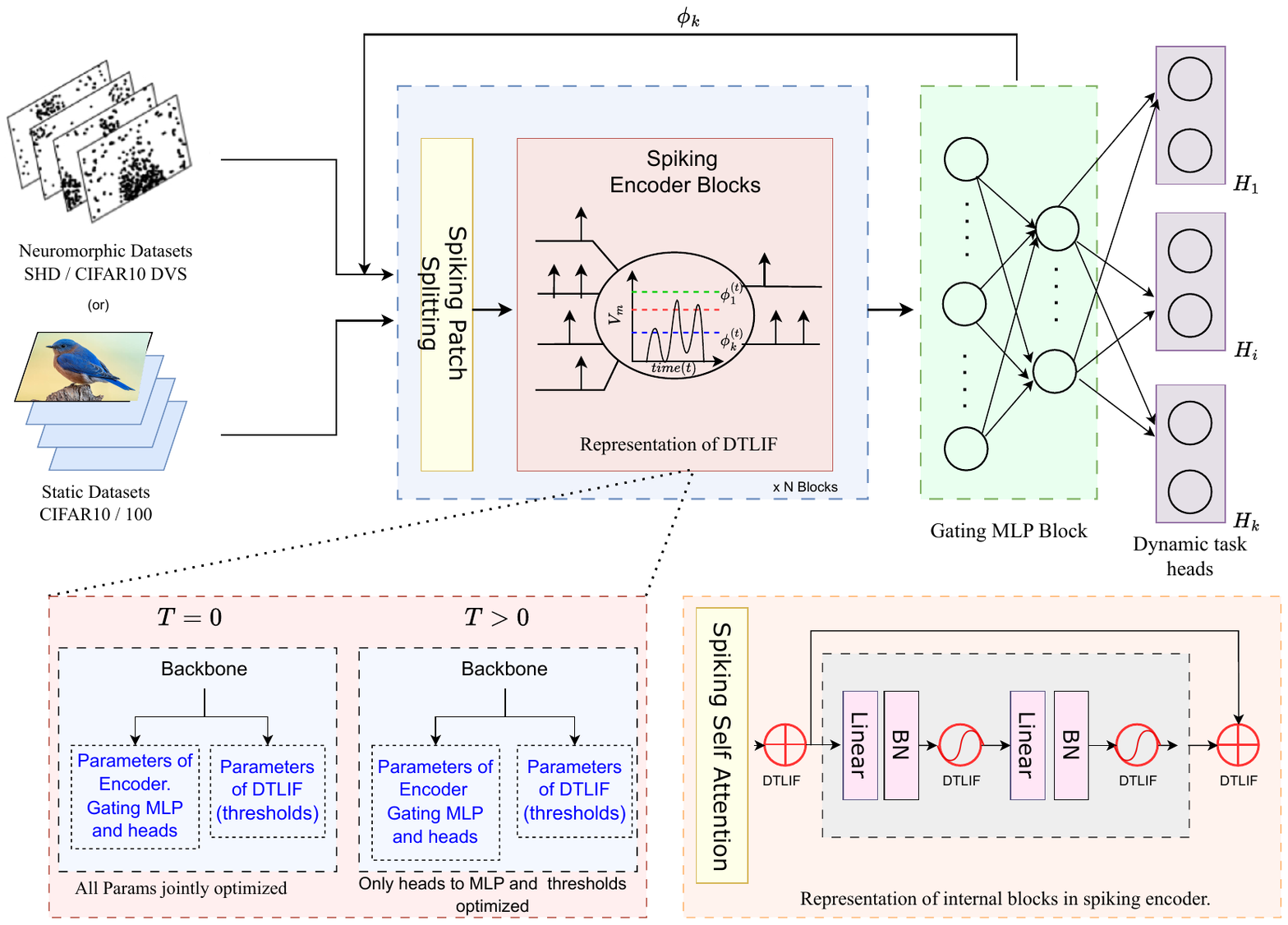}
\caption{The diagram depicts the full architecture and workflow of CATFormer.}
\label{fig: Architecture}
\end{figure*}

\begin{algorithm}[!htbp]
\caption{CATFormer Training Protocol}
\label{alg:training_protocol}
\begin{algorithmic}[1]
\STATE \textbf{Input:} Task sequence $\{\mathcal{T}_k, \mathcal{D}^k\}_{k=1}^T$
\STATE \textbf{Initialize:} Backbone $\theta$, base threshold $\phi_{init} = 0.5$, gating MLP $\mathcal{G}$, buffers $\mathcal{F}_{\text{gate}}, \mathcal{L}_{\text{gate}}$
\FOR{$k = 0$ to $T-1$}
    \STATE Add head $W_k$ (Xavier init), set $\phi^{(k)} \leftarrow \phi_{init}$
    \IF{$k = 0$}
        \STATE Train $\{\theta, \phi^{(0)}, W_0\}$ jointly with $\mathcal{L}_{\text{CE}}$
    \ELSE
        \STATE all previous parameters are frozen; initialize $W_k$ for optimization.
        \STATE Jointly optimize $\{W_k, \phi^{(k)}\}$ , i.e $\min_{\{\phi^{(k)}, W_k\}} \mathbb{E}_{(x,y) \sim \mathcal{D}^k} \left[ \mathcal{L}_{\text{CE}}(W_k \cdot f(x; \theta, \phi^{(k)}), y) \right]$
    \ENDIF
    \STATE Extract features $\mathbf{f}(x)$ using $\phi_{init}$; add $(\mathbf{f}(x), k)$ to buffers \textbf{(local scope not across tasks)}.
    \STATE Train gating MLP $\mathcal{G}$ on accumulated data with $\mathcal{L}_{\text{CE}}$
\ENDFOR
\end{algorithmic}
\end{algorithm}

\begin{algorithm}[!htbp]
\caption{Gated Inference}
\label{alg:inference}
\begin{algorithmic}[1]
\STATE \textbf{Input:} Test sample $x$, trained model $\{\theta, \Phi, \{ W_0 \cdots W_k\}, \mathcal{G}\}$, seen tasks $k$
\STATE \textbf{Output:} Predicted class $\hat{y}$
\IF{$T = 0$}
    \STATE Set thresholds to $\phi^{(0)}$ and reset SNN state
    \STATE Extract features: $\mathbf{f}(x) \leftarrow \text{SpikFormer}(x)$
    \STATE \textbf{return} $\arg\max(W_0 \mathbf{f}(x))$
\ELSE
    \STATE \textbf{Task Prediction:}
    \STATE \quad Set all thresholds to base $\phi_{init}$ and reset SNN state
    \STATE \quad Extract base features: $\mathbf{f}_{\text{base}}(x) \leftarrow \text{SpikFormer}(x)$
    \STATE \quad Predict task: $k^* \leftarrow \arg\max(\mathcal{G}(\mathbf{f}_{\text{base}}(x)))$
    \STATE \textbf{Classification(once the head is selected):}
    \STATE \quad Set thresholds to $\phi^{(k^*)}$ and reset SNN state
    \STATE \quad Extract task-specific features: \\
    $~~~~~~~~~~~~~~~~~~~~~~~~~~\mathbf{f}_{k^*}(x) \leftarrow \text{SpikFormer}(x)$
    \STATE \quad \textbf{return} $\arg\max(W_{k^*} \mathbf{f}_{k^*}(x))$
\ENDIF
\end{algorithmic}
\end{algorithm}

\subsection{Problem Formulation and Notation}
In CIL, the model is exposed to $T$ number of tasks $\{\mathcal{T}_0, \mathcal{T}_1, \ldots, \mathcal{T}_{T-1}\}$ sequentially, each with dataset $\mathcal{D}^k = \{(x_i^k, y_i^k)\}_{i=1}^{N_k}$ and disjoint label sets $\mathcal{Y}_k$ (i.e., $\mathcal{Y}_i \cap \mathcal{Y}_j = \emptyset$ for $i\neq j$). At each task $k$, the model must classify samples over the cumulative label space $\mathcal{Y}_{1:k}$, having access solely to the current task's train data during, and \emph{without the any task oracle or previous task samples} at test time.

\subsection{Context Adaptive Dynamic Threshold Neurons}

\paragraph{Dynamic Threshold LIF Neuron Model:}
To enable task-adaptive spiking dynamics, we extend the standard Leaky Integrate-and-Fire (LIF) neuron with \emph{context adaptive, learnable thresholds}. 
Updation of thresholds :$\tilde V_j^{(t)} = \Bigl(1 - \tfrac{1}{\tau}\Bigr) V_j^{(t-1)} + \tfrac{1}{\tau}\, I_j^{(t)}$ where $\tau$ is the membrane time constant, $I_j^{(t)}$ is the input current.  
$S_j^{(t)} = \Theta\!\bigl(\tilde V_j^{(t)} - \phi_j^{(k)}\bigr)$ 
here $\Theta(\cdot)$ is the Heaviside step function and $S_j^{(t)}$ is the spike output. We use soft reset Mechanism $V_j^{(t)} = \tilde V_j^{(t)} - S_j^{(t)}\, \phi_j^{(k)}$. 
\paragraph{Updation of Dynamic thresholds:} During training on task $k$, the threshold $\phi_c^{(k)}$ is updated via gradient descent: $\phi_j^{(k)} \leftarrow \phi_j^{(k)} - \eta \frac{\partial \mathcal{L}}{\partial \phi_j^{(k)}}$ where $\eta$ is the learning rate and $\mathcal{L}$ is the loss function. 
This mechanism allows each channel to adjust its firing threshold for different tasks, supporting task-adaptive spiking in continual learning.
\subsection{Two-Stage Training Protocol}

Our training protocol balances plasticity and stability by freezing and updating components, using threshold adaptation as the key mechanism to prevent catastrophic forgetting.
Algorithm \ref{alg:training_protocol} outlines the complete training mechanism. 

\subsection{Inference via Dynamic Head Routing}
During inference, we employ our Gated Dynamic Head Selection (G-DHS) mechanism to efficiently route inputs to appropriate task-specific heads.

\paragraph{Gating MLP Architecture.}
The gating network is used as the Gated Dynamic Head Selection (G-DHS), which consists of a two-layer MLP that maps feature embeddings to task predictions:  
    $\mathcal{G}(\mathbf{f}) = \text{Linear}(\text{ReLU}(\text{Linear}(\mathbf{f}))), \text{ where } \mathbf{f} \in \mathbb{R}^D \to \mathbb{R}^{D/4} \to \mathbb{R}^k$. Given an input $x$ and $D$ is the dimension of the feature vector, the inference mechanism is done for this as described by the algorithm \ref{alg:inference}

\section{Results}
\label{sec:results}

\begin{figure*}[!ht]
    \centering
    \includegraphics[width = \textwidth]{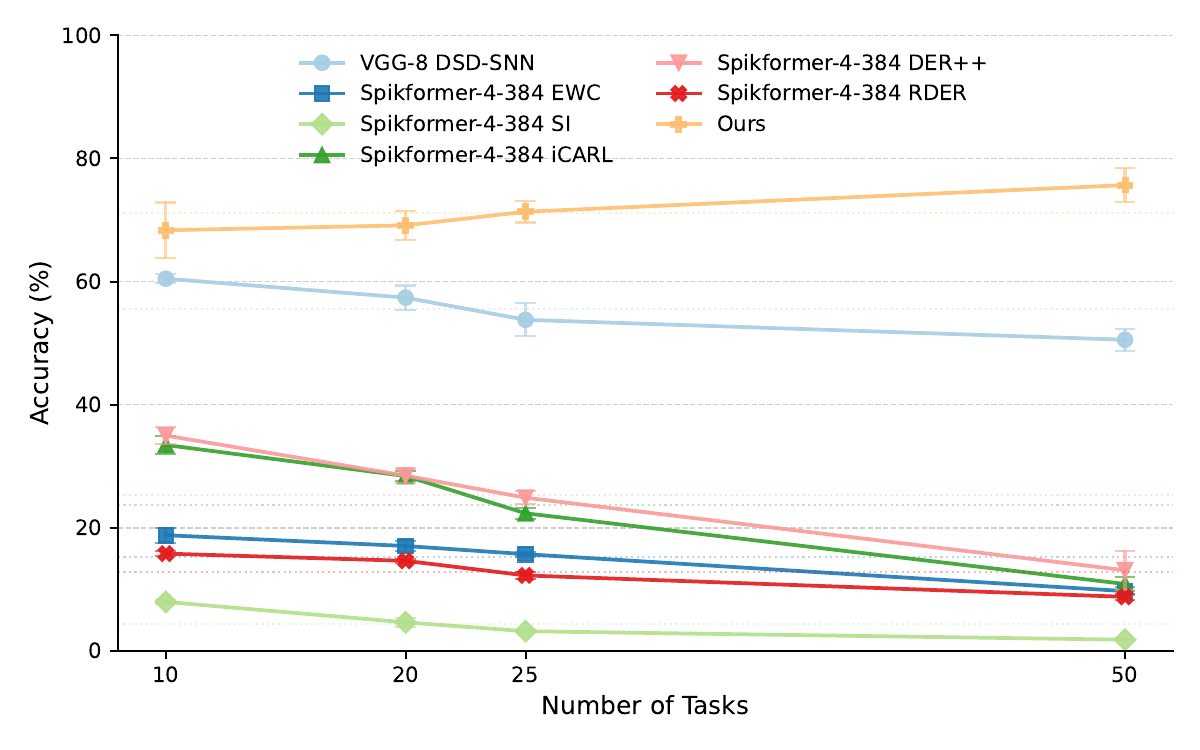}
    \caption{Reverse Forgetting vs Catastrophic Forgetting Trend comparison (No. of Task vs Accuracy( in \%)) of CATFormer against DSD-SNN \cite{dsdsnn} and same SpikFormer on \cite{ewc,zenke2017continual,rebuffi2017icarl,der++,rder}. The dotted line represents the average accuracy across tasks as the number of tasks increases.}
    \label{fig: counter intuitive}
\end{figure*}
\subsection{Dataset and Experimental Setup}
To evaluate the effectiveness of our context-adaptive dynamic threshold mechanism in a spiking Transformer (CATFormer), we conducted extensive experiments across a range of static and neuromorphic datasets. 
We evaluate on  CIFAR-10/100 (10/100 classes) \cite{krizhevsky2009learning_cifar}: standard $32 \times 32$ RGB image benchmarks, tiny-ImageNet (200 classes) \cite{le2015tinyimagenet}: subset of ImageNet with $64 \times 64$ images, CIFAR10-DVS  (10 classes) \cite{cifar10dvs}: a neuromorphic, event-based version of CIFAR-10 captured with Dynamic Vision Sensors, and SHD (20 classes) \cite{cramer2020heidelberg_shd}: a neuromorphic auditory dataset of spiking event sequences preprocessed into fixed-length frames.

\subsection{Experimental Observations}

\subsubsection{Performance in Extended Task Sequences on static datasets}

We compared CATFormer with state-of-the-art methods for class incremental learning in spiking neural networks. For a fair comparison, we evaluated other methods on the same SpikFormer backbone and with respect to reverse forgetting. Further compared Tiny-ImageNet with a non-spiking benchmark with the same SpikFormer backbone.

\begin{table*}[!htbp]
\resizebox{\linewidth}{!}{
\centering
\begin{tabular}{|l|l|cccc|cc|}
\hline
\textbf{Backbone} &\textbf{Methods} & \multicolumn{4}{|c|}{\textbf{Number of Tasks}} & \multicolumn{2}{|c|}{\textbf{Parameters (M)}} \\
\hline
 \multicolumn{2}{|c|}{\textbf{Traditional}} & \multicolumn{4}{|c|}{\textbf{1 Task (Full dataset)}}& \multicolumn{2}{|c|}{\textbf{Total}}\\
\hline
VGG-11&Hybrid& \multicolumn{4}{|c|}{67.87}&\multicolumn{2}{|c|}{9.27}\\
ResNet-19&TET& \multicolumn{4}{|c|}{74.47}&\multicolumn{2}{|c|}{12.63}\\
SpikFormer-4-384&BPTT&\multicolumn{4}{|c|}{77.86}&\multicolumn{2}{|c|}{9.32}\\
\hline
 \multicolumn{2}{|c|}{\textbf{Class incremental Learning}} & \textbf{10 Tasks} & \textbf{20 Tasks} & \textbf{25 Tasks} & \textbf{50 Tasks} & \textbf{Task 0} & \textbf{Task k}  \\
 \hline
VGG-8 &DSD-SNN& 60.47 $\pm$ 0.72 & 57.39 $\pm$ 1.97 & 53.79 $\pm$ 2.67 & 50.55 $\pm$ 1.76 & 14.2 & 14.2 \\
SpikFormer-4-384 & EWC & 18.81 $\pm$ 1.22    &17.06 $\pm$ 0.83    & 15.73 $\pm$ 0.38    & 9.73 $\pm$ 0.62  &    9.32 & 18.64\\
&SI & 7.98 $\pm$ 0.33    & 4.66 $\pm$ 0.74    &  3.22 $\pm$ 0.14 & 1.84 $\pm$ 0.09  & 9.32 & 27.96\\
&iCARL & 33.46 $\pm$ 1.52     & 28.42 $\pm$ 0.77    & 22.37 $\pm$ 0.90    & 10.89 $\pm$ 1.19  & 9.32 & 9.32 \\
&DER++  &  34.99 $\pm$ 1.39      & 28.48 $\pm$1.16     & 24.9 $\pm$1.07    & 13.12 $\pm$ 3.2  &  9.32& 9.32\\
&RDER   & 15.82 $\pm$ 0.36     & 14.65 $\pm$ 0.27     & 12.28 $\pm$ 0.59& 8.8 $\pm$ 0.47  & 11.06 & 11.06 \\
&\textbf{Ours} &\textbf{68.33 $\pm$ 4.51} & \textbf{69.13 $\pm$ 2.36} & \textbf{71.34 $\pm$ 1.75} & \textbf{75.66 $\pm$ 2.72}  & 10.5& \textbf{1.4} \\
\hline
\end{tabular}}
\caption{Comparison of standard CIL accuracy (\%) on Split CIFAR-100 across different task granularities. Reported average test accuracy after all tasks. DSD-SNN results (25/50 tasks)\cite{dsdsnn}; other CIL baselines \cite{ewc, zenke2017continual, rebuffi2017icarl, der++, rder} are evaluated on SpikFormer \cite{zhou2023spikformer}. Task 0 and Task k describe the total number of parameter updates during the $0^{th}$ and $k^{th}$ training tasks, respectively. } 
\label{tab:comparative_results}
\end{table*}
Table \ref{tab:comparative_results} presents our class incremental learning (CIL) performance on Split CIFAR-100. Classical regularisation-based methods like EWC \cite{ewc}, MAS \cite{aljundi2018memory}, and SI \cite{zenke2017continual} suffer from severe catastrophic forgetting. Rehearsal-based approaches such as iCaRL \cite{rebuffi2017icarl} and DER++ \cite{der++} offer better retention, but their memory requirements are often constrained on hardware like the  Lakemont x86 processors and neuromorphic cores in Loihi 2 \cite{Shresthaloihi2}, making even the 2000 samples proposed by ALADE-SNN \cite{ni2025alade} pretty difficult. Among data rehearsal-free SNN baselines, previous state-of-the-art DSD-SNN \cite{dsdsnn} achieves moderate performance but exhibits a consistent forgetting pattern. The accuracy consistently degrades from \textbf{60.47\%} at 10 tasks to \textbf{50.55\%} at 50 tasks. To confirm the trend, we extended the original DSD-SNN repository\footnote{\url{https://github.com/BrainCog-X/Brain-Cog.git}} to generate 25 and 50 task results. On the contrary, \textbf{CATFormer fundamentally breaks this degradation pattern.} Our model not only surpasses the classical CIL baselines by substantial margins but also demonstrates the unprecedented behaviour of \emph{improving} accuracy with increasing tasks from \textbf{68.33\%} at 10 tasks to \textbf{75.66\%} at 50 tasks. This is a counter-intuitive `reverse forgetting' trend, as illustrated in Figure \ref{fig: counter intuitive}. This effect stems from our dynamic threshold adaptation with the gating mechanism, which optimises neuronal firing for new tasks without overwriting prior knowledge. This trend toward long task sequences is crucial for real-world applications, as highlighted in robotics and embodied AI research \cite{lesort2020continual,hajizada2022interactive}, which involve continuous adaptation and data drift. We evaluated our model performance on the 10-class count CIFAR-10 dataset. The comparative results in Table \ref{tab:neuromorphic_results}, with the best-performing model, DSD-SNN \cite{shen2024efficient}, serving as the baseline, demonstrate CATFormer's superior performance on Split CIFAR-10, achieving an accuracy of \textbf{89.29\%} for the 5-task split. 

\paragraph{Comparison of parameter updates at $k^{th}$ task}
In terms of model size, our proposed architecture is parameter-efficient. The base SpikFormer \cite{zhou2023spikformer} has a base size of 9.32M parameters, while our model has a base size of 10.5M, of which 1.2M is attributed to the routing mechanism. Although there is a minimal increase in the number of parameters, it is comparable to other methods using the same SpikFormer backbone. For instance, a task 0 CATFormer is comparable to most of the current training paradigms presented in Table \ref{tab:comparative_results}. Our approach becomes significantly parameter-efficient once we move to tasks $T^k$, where $k>0$, where we approximately train only 1.4 million parameters, while the other model actually updates the entire model's parameters and also utilises a rehearsal buffer in some cases (which can also lead to unwanted privacy breaches). The total parameter count for our model scales efficiently with the term count ($\phi^k$), where approximately 16,032 thresholds per task are stored, requiring no more than 64.2 KB of memory when FP32 is used for storage. This can eventually be reduced to FP16, since we hardly need many decimal places, indicating that only a small number of task-specific parameters are added as new tasks are learned, while achieving state-of-the-art performance.

\subsubsection{Performance on tiny-ImageNet Dataset}
We further evaluate CATFormer's performance on the 200-class count Tiny-ImageNet dataset. Since CIFAR-100 allows up to 50 task training, we test the efficacy of our model on 100 tasks that were not possible earlier. Due to the unavailability of a previous similar CIL implementation of SNNs on ImageNet, we compare performance to an ANN baseline \cite{liu2025cvpr_ssm_tinyimagenet}, which features a non-transformer backbone \cite{mamba} on \cite{moe-mose_cvpr24}. We observe \textbf{8.45\%} improvement over the baselines (Table \ref{tab:neuromorphic_results}). Hence, CATFormer achieves better overall performance than non-spiking models across a larger number of tasks.


\subsubsection{Performance on Spiking Datasets}
Neuromorphic datasets pose challenges due to their spatiotemporal dynamics and event-driven inputs, making them ideal testbeds for our dynamic threshold hypothesis. Table \ref{tab:neuromorphic_results} demonstrates CATFormer's efficacy on SHD (with only timesteps of 16) and CIFAR10-DVS benchmarks, achieving \textbf{84.48\%}/\textbf{87.85\%} and \textbf{83.21\%}/\textbf{87.14\%} for 5/10 and 2/5 task splits, respectively. Notably, unlike our other computer vision evaluation datasets, namely CIFAR-10/100, CIFAR10-DVS, SHD is an audio-based classification dataset. We observe that, even without any data rehearsal strategy, CATFormer's \textbf{83.21\%} performance on CIFAR10-DVS (2 tasks) closely matches the rehearsal augmented performance of ALADE-SNNs \cite{ni2025alade}, 83.5\%. CATFormer thereby establishes a new standard for data rehearsal-free neuromorphic continual learning.

The consistent performance gains across neuromorphic datasets validate that our context-adaptive threshold mechanism naturally aligns with the temporal processing characteristics inherent to spiking neural networks, enabling more effective utilisation of the temporal dimension for task differentiation. This neuromorphic compatibility represents a significant advancement, as these datasets remain underexplored in data rehearsal-free continual learning scenarios.

\begin{table}[!htbp]
\centering
\begin{tabular}{|l|c|c|c|}
\hline
\textbf{Dataset} & \textbf{Task} & \textbf{Model} & \textbf{Accuracy} \\
\hline
CIFAR10 & 5 & SA-SNN+EWC & 80.39 $\pm$ 1.84\\
& & \textbf{CATFormer} & \textbf{89.29 $\pm$ 2.53} \\
\hline
CIFAR10 & 2 & DSD-SNN & 80.90 $\pm$ 1.20 \\
-DVS& & \textbf{CATFormer} &  \textbf{83.21 $\pm$ 2.33} \\
& 5 & DSD-SNN & 76.57 $\pm$ 0.96 \\
& & \textbf{CATFormer} &  \textbf{87.14 $\pm$ 2.78} \\
\hline
SHD & 5  & DSD-SNN & 82.56 $\pm$ 1.15 \\
(T=16)& & \textbf{CATFormer} &  \textbf{84.48 $\pm$ 1.62} \\
& 10 & DSD-SNN & 80.47 $\pm$ 1.03 \\
& & \textbf{CATFormer} &  \textbf{87.85 $\pm$ 1.20} \\
\hline
Tiny Im-Net & \textbf{100} & S6MOD$\textbf{*}$   & 40.11 $\pm$ 0.26 \\
& & \textbf{CATFormer} &  \textbf{48.56 $\pm$ 0.81} \\
\hline
\end{tabular}
\caption{Task-wise accuracy (\%) of CATFormer with respect to the state-of-the-art SNNs on static and neuromorphic datasets. Methods marked with \textbf{*} denote online continual learning approaches.}
\label{tab:neuromorphic_results}
\end{table}

\subsection{Ablation Studies}

We conducted targeted ablation experiments on Split CIFAR-10 (5 tasks, 2 classes per task) to isolate the impact of each core component in CATFormer. The ablated variants and their average accuracy (in \%) after all tasks are described in Table \ref {tab:ablation_cifar10}. The \textbf{Fixed Threshold} variant, where all neurons use their initial firing threshold for all tasks, leads to pronounced catastrophic forgetting, with accuracy dropping to 42.87\%. In this configuration, performance on the first task is reasonable, but subsequent tasks trigger a consistent 15–18\% degradation after each increment, mirroring classical forgetting patterns in SNNs without adaptive mechanisms. Hence, we observe that, when studying catastrophic forgetting in brain-inspired SNNs, it is important to pay close attention to the role of spiking thresholds, unlike previous studies \cite{ni2025alade, dsdsnn, shen2024efficient}. Biologically, such dynamic threshold behavior can potentially be mediated by neuromodulation \cite{tsuda2026neuromodulators,oh2015increased,liu2022biologically,xu2005activity}.

\begin{table}[!htbp]
\centering
\begin{tabular}{|l|c|c|}
\hline
\textbf{Ablation Variant} &\textbf{Accuracy}  &\textbf{Acc. on task $\mathbf0$}\\
\hline
Fixed Threshold           & 42.87$\pm$1.26 & 72.59$\pm$1.86 \\
SpikIdentityFormer        & 59.38$\pm$0.98 & 70.62$\pm$1.75\\
Random Identity Former    & 53.17$\pm$2.13 & 62.43$\pm$0.99\\
FFN Frozen                & 63.24$\pm$1.78 & 72.17$\pm$1.59\\
\hline
\textbf{CATFormer} & \textbf{89.29$\pm$2.53} & \textbf{93.87$\pm$0.45} \\
\hline
\end{tabular}
\caption{Ablation study of CATFormer evaluated on CIFAR-10 (5 tasks).}
\label{tab:ablation_cifar10}
\end{table}

Recent works on vision transformer \cite{yu2022metaformeractuallyneedvision, metaformer2} show that its performance is not significantly impacted by the choice of token mixers. In fact, they show that even after removing all vanilla attention token mixers, the ViT achieves good baseline performance. Driven by these observations, we evaluate the performance of our spiking vision transformer (CATFormer) on the CIFAR-10 (5 tasks) platform. As part of this, we design \textbf{SpikIdentityFormer} by either replacing all spike attention with identity mapping in CATFormer (\textbf{SpikIdentityFormer}) or replacing spike attention with uniformly distributed random numbers in CATFormer (\textbf{Random Identity Former}). Both the \textbf{SpikIdentityFormer} (removing all attention) and \textbf{Random Identity Former} (disrupting transformer blocks) yield similar, markedly reduced accuracy, confirming that structured feature transformation is crucial for robust continual learning. Significantly, a \textbf{Random Identity Former} performs lower (53.17\%) than a \textbf{SpikIdentityFormer} (59.38\%). Moreover, freezing the feed-forward network (\textbf{FFN Frozen}) provides only moderate improvement (63.24\%), indicating that adaptive intermediate representations are also necessary for effective knowledge retention over time. Compared to only FFN learning in \textbf{SpikIdentityFormer}, a learnable token mixer in \textbf{FFN Frozen} (with FFN frozen) performs better by 3.86\%. A similar trend is observed in first-task learning. These results demonstrate that both the transformer’s structured feature extraction and, critically, its dynamic, task-specific threshold modulation are essential for effective, data-rehearsal-free continual learning, thereby justifying our design choices. 
Overall, CATFormer surpasses prior rehearsal-free methods on static and neuromorphic benchmarks, achieving stable or improved accuracy as the task count increases. This makes it well-suited for continual learning on embedded, memory-constrained robotic and autonomous systems that must adapt to dynamic, unpredictable environments. Unlike conventional rehearsal-based methods, which face prohibitive storage, energy, privacy, and bandwidth constraints, CATFormer almost eliminates the need for memory buffers by leveraging dynamic, task-specific neuronal plasticity. This biologically inspired and hardware-efficient design aligns with practical constraints highlighted in robotics and embodied AI research \cite{lesort2020continual, hajizada2022interactive}, enabling robust lifelong continual learning in real-world deployment.

\section{Discussion}
\label{sec:discussion}
CATFormer demonstrates that biologically inspired dynamic threshold adaptation enables rehearsal-free continual learning in spiking neural networks, maintaining or improving accuracy across up to 50 tasks. This is especially relevant for robotics and physical AI, where limited onboard memory makes storing replay buffers impractical—for instance, retaining 2,000 CIFAR-100 images for rehearsal consumes \cite{ni2025alade} approximately 25–30MB, a significant overhead for resource-constrained hardware. Moreover, continual network growth or pruning introduces complexity and unstable resource demands, hindering deployment on embedded or neuromorphic platforms. By leveraging intrinsic neuronal modulation through task-specific dynamic thresholds, CATFormer provides a memory and computation-efficient solution that avoids these pitfalls while sustaining robust, scalable continual learning. Future work should explore adaptive threshold learning in streaming, non-stationary environments for \textit{lifelong learning} and investigate direct hardware implementations on Loihi 2 \cite{Shresthaloihi2} to accelerate the real-world deployment of lifelong SNN agents.

\section{Acknowledgments}

This work was supported by the Faculty Startup Grant from IIT Guwahati. 
\bibliographystyle{unsrt}
\bibliography{references}

@article{threeparadiagms,
  title={Three scenarios for continual learning},
  author={Van de Ven, Gido M and Tolias, Andreas S},
  journal={arXiv preprint arXiv:1904.07734},
  year={2019}
}

@inproceedings{cvpr25taskspecificadapters,
  title={Dynamic integration of task-specific adapters for class incremental learning},
  author={Li, Jiashuo and Wang, Shaokun and Qian, Bo and He, Yuhang and Wei, Xing and Wang, Qiang and Gong, Yihong},
  booktitle={Proceedings of the Computer Vision and Pattern Recognition Conference},
  pages={30545--30555},
  year={2025}
}

@article{hao2020biologically,
  title={A biologically plausible supervised learning method for spiking neural networks using the symmetric STDP rule},
  author={Hao, Yunzhe and Huang, Xuhui and Dong, Meng and Xu, Bo},
  journal={Neural Networks},
  volume={121},
  pages={387--395},
  year={2020},
  publisher={Elsevier}
}

@article{kim2021spiking,
  title={Spiking neural network (SNN) with memristor synapses having non-linear weight update},
  author={Kim, Taeyoon and Hu, Suman and Kim, Jaewook and Kwak, Joon Young and Park, Jongkil and Lee, Suyoun and Kim, Inho and Park, Jong-Keuk and Jeong, YeonJoo},
  journal={Frontiers in computational neuroscience},
  volume={15},
  pages={646125},
  year={2021},
  publisher={Frontiers Media SA}
}

@article{shaban2021adaptive,
  title={An adaptive threshold neuron for recurrent spiking neural networks with nanodevice hardware implementation},
  author={Shaban, Ahmed and Bezugam, Sai Sukruth and Suri, Manan},
  journal={Nature Communications},
  volume={12},
  number={1},
  pages={4234},
  year={2021},
  publisher={Nature Publishing Group UK London}
}

@inproceedings{moe-mose_cvpr24,
  title={Orchestrate latent expertise: Advancing online continual learning with multi-level supervision and reverse self-distillation},
  author={Yan, Hongwei and Wang, Liyuan and Ma, Kaisheng and Zhong, Yi},
  booktitle={Proceedings of the IEEE/CVF Conference on Computer Vision and Pattern Recognition},
  pages={23670--23680},
  year={2024}
}

@article{cifar10dvs,
  title={Cifar10-dvs: an event-stream dataset for object classification},
  author={Li, Hongmin and Liu, Hanchao and Ji, Xiangyang and Li, Guoqi and Shi, Luping},
  journal={Frontiers in neuroscience},
  volume={11},
  pages={244131},
  year={2017},
  publisher={Frontiers}
}

@inproceedings{mamba,
  title={Mamba: Linear-time sequence modeling with selective state spaces},
  author={Gu, Albert and Dao, Tri},
  booktitle={First conference on language modeling},
  year={2024}
}

@article{cramer2020heidelberg_shd,
  title={The heidelberg spiking data sets for the systematic evaluation of spiking neural networks},
  author={Cramer, Benjamin and Stradmann, Yannik and Schemmel, Johannes and Zenke, Friedemann},
  journal={IEEE Transactions on Neural Networks and Learning Systems},
  volume={33},
  number={7},
  pages={2744--2757},
  year={2020},
  publisher={IEEE}
}

@article{krizhevsky2009learning_cifar,
  title={Learning multiple layers of features from tiny images},
  author={Krizhevsky, Alex and Hinton, Geoffrey and others},
  year={2009},
  publisher={Toronto, ON, Canada}
}

@article{le2015tinyimagenet,
  title={Tiny imagenet visual recognition challenge},
  author={Le, Yann and Yang, Xuan and others},
  journal={CS 231N},
  volume={7},
  number={7},
  pages={3},
  year={2015}
}

@inproceedings{liu2025cvpr_ssm_tinyimagenet,
  title={Enhancing online continual learning with plug-and-play state space model and class-conditional mixture of discretization},
  author={Liu, Sihao and Yang, Yibo and Li, Xiaojie and Clifton, David A and Ghanem, Bernard},
  booktitle={Proceedings of the Computer Vision and Pattern Recognition Conference},
  pages={20502--20511},
  year={2025}
}

@inproceedings{he2025cllora,
  title={CL-LoRA: Continual low-rank adaptation for rehearsal-free class-incremental learning},
  author={He, Jiangpeng and Duan, Zhihao and Zhu, Fengqing},
  booktitle={Proceedings of the Computer Vision and Pattern Recognition Conference},
  pages={30534--30544},
  year={2025}
}

@article{hasselmo1995cholinergic,
  title={Cholinergic modulation of activity-dependent synaptic plasticity in the piriform cortex and associative memory function in a network biophysical simulation},
  author={Hasselmo, Michael E and Barkai, Edi},
  journal={Journal of Neuroscience},
  volume={15},
  number={10},
  pages={6592--6604},
  year={1995},
  publisher={Society for Neuroscience}
}

@article{miller2001integrative_pnas,
  title={An integrative theory of prefrontal cortex function},
  author={Miller, Earl K and Cohen, Jonathan D},
  journal={Annual review of neuroscience},
  volume={24},
  number={1},
  pages={167--202},
  year={2001},
  publisher={Annual Reviews 4139 El Camino Way, PO Box 10139, Palo Alto, CA 94303-0139, USA}
}

@article{engel2001dynamic_pnas,
  title={Dynamic predictions: oscillations and synchrony in top--down processing},
  author={Engel, Andreas K and Fries, Pascal and Singer, Wolf},
  journal={Nature Reviews Neuroscience},
  volume={2},
  number={10},
  pages={704--716},
  year={2001},
  publisher={Nature Publishing Group UK London}
}

@article{masse2018alleviatin,
  title={Alleviating catastrophic forgetting using context-dependent gating and synaptic stabilization},
  author={Masse, Nicolas Y and Grant, Gregory D and Freedman, David J},
  journal={Proceedings of the National Academy of Sciences},
  volume={115},
  number={44},
  pages={E10467--E10475},
  year={2018},
  publisher={National Academy of Sciences}
}

@article{dosovitskiy2020image,
  title={An image is worth 16x16 words: Transformers for image recognition at scale},
  author={Dosovitskiy, Alexey and Beyer, Lucas and Kolesnikov, Alexander and Weissenborn, Dirk and Zhai, Xiaohua and Unterthiner, Thomas and Dehghani, Mostafa and Minderer, Matthias and Heigold, Georg and Gelly, Sylvain and others},
  journal={arXiv preprint arXiv:2010.11929},
  year={2020}
}

@article{wang2024comprehensive,
  title={A comprehensive survey of continual learning: Theory, method and application},
  author={Wang, Liyuan and Zhang, Xingxing and Su, Hang and Zhu, Jun},
  journal={IEEE transactions on pattern analysis and machine intelligence},
  volume={46},
  number={8},
  pages={5362--5383},
  year={2024},
  publisher={IEEE}
}

@article{chaudhry2018efficient,
  title={Efficient lifelong learning with a-gem},
  author={Chaudhry, Arslan and Ranzato, Marc'Aurelio and Rohrbach, Marcus and Elhoseiny, Mohamed},
  journal={arXiv preprint arXiv:1812.00420},
  year={2018}
}

@inproceedings{selfexpansioncvpr,
  title={Self-expansion of pre-trained models with mixture of adapters for continual learning},
  author={Wang, Huiyi and Lu, Haodong and Yao, Lina and Gong, Dong},
  booktitle={Proceedings of the Computer Vision and Pattern Recognition Conference},
  pages={10087--10098},
  year={2025}
}

@inproceedings{representationexpansion,
  title={Self-sustaining representation expansion for non-exemplar class-incremental learning},
  author={Zhu, Kai and Zhai, Wei and Cao, Yang and Luo, Jiebo and Zha, Zheng-Jun},
  booktitle={Proceedings of the IEEE/CVF conference on computer vision and pattern recognition},
  pages={9296--9305},
  year={2022}
}

@article{hasselmo2006role,
  title={The role of acetylcholine in learning and memory},
  author={Hasselmo, Michael E},
  journal={Current opinion in neurobiology},
  volume={16},
  number={6},
  pages={710--715},
  year={2006},
  publisher={Elsevier}
}

@inproceedings{cvpr25mem_eff_reh,
  title={Online task-free continual learning via dynamic expansionable memory distribution},
  author={Ye, Fei and Bors, Adrian G},
  booktitle={Proceedings of the Computer Vision and Pattern Recognition Conference},
  pages={20512--20522},
  year={2025}
}

@inproceedings{liang2024inflora,
  title={Inflora: Interference-free low-rank adaptation for continual learning},
  author={Liang, Yan-Shuo and Li, Wu-Jun},
  booktitle={Proceedings of the IEEE/CVF Conference on Computer Vision and Pattern Recognition},
  pages={23638--23647},
  year={2024}
}

@inproceedings{
zhou2023spikformer,
title={Spikformer: When Spiking Neural Network Meets Transformer },
author={Zhaokun Zhou and Yuesheng Zhu and Chao He and Yaowei Wang and Shuicheng YAN and Yonghong Tian and Li Yuan},
booktitle={The Eleventh International Conference on Learning Representations },
year={2023},
url={https://openreview.net/forum?id=frE4fUwz_h}
}

@article{dsdsnn,
  title={Enhancing efficient continual learning with dynamic structure development of spiking neural networks},
  author={Han, Bing and Zhao, Feifei and Zeng, Yi and Pan, Wenxuan and Shen, Guobin},
  journal={arXiv preprint arXiv:2308.04749},
  year={2023}
}

@inproceedings{ni2025alade,
  title={Alade-snn: Adaptive logit alignment in dynamically expandable spiking neural networks for class incremental learning},
  author={Ni, Wenyao and Shen, Jiangrong and Xu, Qi and Tang, Huajin},
  booktitle={Proceedings of the AAAI Conference on Artificial Intelligence},
  volume={39},
  pages={19712--19720},
  year={2025}
}

@article{
hammouamri2022mitigating,
title={Mitigating Catastrophic Forgetting in Spiking Neural Networks through Threshold Modulation},
author={Ilyass Hammouamri and Timoth{\'e}e Masquelier and Dennis George Wilson},
journal={Transactions on Machine Learning Research},
issn={2835-8856},
year={2022},
url={https://openreview.net/forum?id=15SoThZmtU},
note={}
}

@inproceedings{shen2024efficient,
  title={Efficient spiking neural networks with sparse selective activation for continual learning},
  author={Shen, Jiangrong and Ni, Wenyao and Xu, Qi and Tang, Huajin},
  booktitle={Proceedings of the AAAI Conference on Artificial Intelligence},
  volume={38},
  pages={611--619},
  year={2024}
}

@article{ewc,
  title={Overcoming catastrophic forgetting in neural networks},
  author={Kirkpatrick, James and Pascanu, Razvan and Rabinowitz, Neil and Veness, Joel and Desjardins, Guillaume and Rusu, Andrei A and Milan, Kieran and Quan, John and Ramalho, Tiago and Grabska-Barwinska, Agnieszka and others},
  journal={Proceedings of the national academy of sciences},
  volume={114},
  number={13},
  pages={3521--3526},
  year={2017},
  publisher={National Academy of Sciences}
}

@inproceedings{aljundi2018memory,
  title={Memory aware synapses: Learning what (not) to forget},
  author={Aljundi, Rahaf and Babiloni, Francesca and Elhoseiny, Mohamed and Rohrbach, Marcus and Tuytelaars, Tinne},
  booktitle={Proceedings of the European conference on computer vision (ECCV)},
  pages={139--154},
  year={2018}
}

@inproceedings{zenke2017continual,
  title={Continual learning through synaptic intelligence},
  author={Zenke, Friedemann and Poole, Ben and Ganguli, Surya},
  booktitle={International conference on machine learning},
  pages={3987--3995},
  year={2017},
  organization={Pmlr}
}

@inproceedings{rebuffi2017icarl,
  title={icarl: Incremental classifier and representation learning},
  author={Rebuffi, Sylvestre-Alvise and Kolesnikov, Alexander and Sperl, Georg and Lampert, Christoph H},
  booktitle={Proceedings of the IEEE conference on Computer Vision and Pattern Recognition},
  pages={2001--2010},
  year={2017}
}

@inproceedings{hajizada2022interactive,
  title={Interactive continual learning for robots: a neuromorphic approach},
  author={Hajizada, Elvin and Berggold, Patrick and Iacono, Massimiliano and Glover, Arren and Sandamirskaya, Yulia},
  booktitle={Proceedings of the international conference on neuromorphic systems 2022},
  pages={1--10},
  year={2022}
}

@article{rder,
  title={Relational experience replay: Continual learning by adaptively tuning task-wise relationship},
  author={Wang, Quanziang and Wang, Renzhen and Li, Yuexiang and Wei, Dong and Wang, Hong and Ma, Kai and Zheng, Yefeng and Meng, Deyu},
  journal={IEEE Transactions on Multimedia},
  volume={26},
  pages={9683--9698},
  year={2024},
  publisher={IEEE}
}

@article{der++,
  title={Dark experience for general continual learning: a strong, simple baseline},
  author={Buzzega, Pietro and Boschini, Matteo and Porrello, Angelo and Abati, Davide and Calderara, Simone},
  journal={Advances in neural information processing systems},
  volume={33},
  pages={15920--15930},
  year={2020}
}

@article{lin2025onlinecontinuallearningspiking,
  title={Online Continual Learning via Spiking Neural Networks with Sleep Enhanced Latent Replay},
  author={Lin, Erliang and Luo, Wenbin and Jia, Wei and Chen, Yu and Yang, Shaofu},
  journal={arXiv preprint arXiv:2507.02901},
  year={2025}
}

@inproceedings{Shresthaloihi2,
  title={Efficient video and audio processing with loihi 2},
  author={Shrestha, Sumit Bam and Timcheck, Jonathan and Frady, Paxon and Campos-Macias, Leobardo and Davies, Mike},
  booktitle={ICASSP 2024-2024 IEEE International Conference on Acoustics, Speech and Signal Processing (ICASSP)},
  pages={13481--13485},
  year={2024},
  organization={IEEE}
}

@article{ding2022biologically,
  title={Biologically inspired dynamic thresholds for spiking neural networks},
  author={Ding, Jianchuan and Dong, Bo and Heide, Felix and Ding, Yufei and Zhou, Yunduo and Yin, Baocai and Yang, Xin},
  journal={Advances in neural information processing systems},
  volume={35},
  pages={6090--6103},
  year={2022}
}

@article{fernando2017pathnet,
  title={Pathnet: Evolution channels gradient descent in super neural networks},
  author={Fernando, Chrisantha and Banarse, Dylan and Blundell, Charles and Zwols, Yori and Ha, David and Rusu, Andrei A and Pritzel, Alexander and Wierstra, Daan},
  journal={arXiv preprint arXiv:1701.08734},
  year={2017}
}

@article{rusu2016progressive,
  title={Progressive neural networks},
  author={Rusu, Andrei A and Rabinowitz, Neil C and Desjardins, Guillaume and Soyer, Hubert and Kirkpatrick, James and Kavukcuoglu, Koray and Pascanu, Razvan and Hadsell, Raia},
  journal={arXiv preprint arXiv:1606.04671},
  year={2016}
}

@article{liu2022biologically,
  title={Biologically-plausible backpropagation through arbitrary timespans via local neuromodulators},
  author={Liu, Yuhan Helena and Smith, Stephen and Mihalas, Stefan and Shea-Brown, Eric and S{\"u}mb{\"u}l, Uygar},
  journal={Advances in Neural Information Processing Systems},
  volume={35},
  pages={17528--17542},
  year={2022}
}

@article{metaformer2,
  title={Metaformer baselines for vision},
  author={Yu, Weihao and Si, Chenyang and Zhou, Pan and Luo, Mi and Zhou, Yichen and Feng, Jiashi and Yan, Shuicheng and Wang, Xinchao},
  journal={IEEE Transactions on Pattern Analysis and Machine Intelligence},
  volume={46},
  number={2},
  pages={896--912},
  year={2023},
  publisher={IEEE}
}

@article{grossberg2017acetylcholine,
  title={Acetylcholine neuromodulation in normal and abnormal learning and memory: vigilance control in waking, sleep, autism, amnesia and Alzheimer’s disease},
  author={Grossberg, Stephen},
  journal={Frontiers in neural circuits},
  volume={11},
  pages={82},
  year={2017},
  publisher={Frontiers Media SA}
}

@inproceedings{yu2022metaformeractuallyneedvision,
    author    = {Yu, Weihao and Luo, Mi and Zhou, Pan and Si, Chenyang and Zhou, Yichen and Wang, Xinchao and Feng, Jiashi and Yan, Shuicheng},
    title     = {MetaFormer Is Actually What You Need for Vision},
    booktitle = {Proceedings of the IEEE/CVF Conference on Computer Vision and Pattern Recognition (CVPR)},
    month     = {June},
    year      = {2022},
    pages     = {10819-10829}
}

@inproceedings{wei2023temporal,
  title={Temporal-coded spiking neural networks with dynamic firing threshold: Learning with event-driven backpropagation},
  author={Wei, Wenjie and Zhang, Malu and Qu, Hong and Belatreche, Ammar and Zhang, Jian and Chen, Hong},
  booktitle={Proceedings of the IEEE/CVF international conference on computer vision},
  pages={10552--10562},
  year={2023}
}

@article{huang2016adaptive,
  title={Adaptive spike threshold enables robust and temporally precise neuronal encoding},
  author={Huang, Chao and Resnik, Andrey and Celikel, Tansu and Englitz, Bernhard},
  journal={PLoS computational biology},
  volume={12},
  number={6},
  pages={e1004984},
  year={2016},
  publisher={Public Library of Science San Francisco, CA USA}
}

@article{farmer2012learning,
  title={Learning-dependent plasticity of hippocampal CA1 pyramidal neuron postburst afterhyperpolarizations and increased excitability after inhibitory avoidance learning depend upon basolateral amygdala inputs},
  author={Farmer, George E and Thompson, Lucien T},
  journal={Hippocampus},
  volume={22},
  number={8},
  pages={1703--1719},
  year={2012},
  publisher={Wiley Online Library}
}

@article{beaulieu2020learning,
  title={Learning to continually learn},
  author={Beaulieu, Shawn and Frati, Lapo and Miconi, Thomas and Lehman, Joel and Stanley, Kenneth O and Clune, Jeff and Cheney, Nick},
  journal={arXiv preprint arXiv:2002.09571},
  year={2020}
}

@article{lesort2020continual,
  title={Continual learning for robotics: Definition, framework, learning strategies, opportunities and challenges},
  author={Lesort, Timoth{\'e}e and Lomonaco, Vincenzo and Stoian, Andrei and Maltoni, Davide and Filliat, David and D{\'\i}az-Rodr{\'\i}guez, Natalia},
  journal={Information fusion},
  volume={58},
  pages={52--68},
  year={2020},
  publisher={Elsevier}
}

@article{tsuda2026neuromodulators,
  title={Neuromodulators Generate Multiple Context-Relevant Behaviors in Recurrent Neural Networks},
  author={Tsuda, Ben and Pate, Stefan C and Tye, Kay M and Siegelmann, Hava T and Sejnowski, Terrence J},
  journal={Neural Computation},
  pages={1--36},
  year={2026},
  publisher={MIT Press 255 Main Street, 9th Floor, Cambridge, Massachusetts 02142, USA~…}
}

@article{oh2015increased,
  title={Increased excitability of both principal neurons and interneurons during associative learning},
  author={Oh, M Matthew and Disterhoft, John F},
  journal={The Neuroscientist},
  volume={21},
  number={4},
  pages={372--384},
  year={2015},
  publisher={SAGE Publications Sage CA: Los Angeles, CA}
}

@article{xu2005activity,
  title={Activity-dependent long-term potentiation of intrinsic excitability in hippocampal CA1 pyramidal neurons},
  author={Xu, Jun and Kang, Ning and Jiang, Li and Nedergaard, Maiken and Kang, Jian},
  journal={Journal of Neuroscience},
  volume={25},
  number={7},
  pages={1750--1760},
  year={2005},
  publisher={Society for Neuroscience}
}

\end{document}